\documentclass[sigconf, nonacm, preprint]{acmart}

\renewcommand\footnotetextcopyrightpermission[1]{}
\usepackage{graphicx}
\usepackage{booktabs}  
\usepackage{tabularx}
\usepackage{multirow}  
\usepackage{subcaption} 
\usepackage{pifont}    
\usepackage{float}

\AtBeginDocument{%
  }

\begin{document}

\title{DetailVerifyBench: A Benchmark for Dense Hallucination Localization in Long Image Captions}

\author{Xinran Wang}
\authornote{Both authors contributed equally to this research.}
\email{wangxr@bupt.edu.cn}
\author{Yuxuan Zhang}
\authornotemark[1]
\email{zyx_hhnkh@bupt.edu.cn}
\affiliation{%
  \institution{Beijing University of Posts and Telecommunications}
  \city{Beijing}
  \country{China}
}

\author{Xiao Zhang}
\email{xiaozhang@bupt.edu.cn}
\author{Haolong Yan}
\email{yanhaolong@bupt.edu.cn}
\affiliation{%
  \institution{Beijing University of Posts and Telecommunications}
  \city{Beijing}
  \country{China}
}

\author{Muxi Diao}
\email{dmx@bupt.edu.cn}
\author{Songyu Xu}
\email{xusongyu@bupt.edu.cn}
\affiliation{%
  \institution{Beijing University of Posts and Telecommunications}
  \city{Beijing}
  \country{China}
}

\author{Zhonghao Yan}
\email{zhonghao.yan@bupt.edu.cn}
\author{Hongbing Li}
\email{hbl@bupt.edu.cn}
\affiliation{%
  \institution{Beijing University of Posts and Telecommunications}
  \city{Beijing}
  \country{China}}

\author{Kongming Liang}
\authornote{Corresponding author.}
\email{liangkongming@bupt.edu.cn}
\affiliation{%
  \institution{Beijing University of Posts and Telecommunications}
  \city{Beijing}
  \country{China}}

\author{Zhanyu Ma}
\email{mazhanyu@bupt.edu.cn}
\affiliation{%
  \institution{Beijing University of Posts and Telecommunications}
  \city{Beijing}
  \country{China}}


\begin{abstract}
  Accurately detecting and localizing hallucinations is a critical task for ensuring high reliability of image captions. In the era of Multimodal Large Language Models (MLLMs), captions have evolved from brief sentences into comprehensive narratives, often spanning hundreds of words. This shift exponentially increases the challenge: models must now pinpoint specific erroneous spans or words within extensive contexts, rather than merely flag response-level inconsistencies. However, existing benchmarks lack the fine granularity and domain diversity required to evaluate this capability. To bridge this gap, we introduce DetailVerifyBench, a rigorous benchmark comprising 1,000 high-quality images across five distinct domains. With an average caption length of over 200 words and dense, token-level annotations of multiple hallucination types, it stands as the most challenging benchmark for precise hallucination localization in the field of long image captioning to date. Our benchmark is available at \url{https://zyx-hhnkh.github.io/DetailVerifyBench/}.
\end{abstract}



\begin{teaserfigure}
  \includegraphics[width=\textwidth]{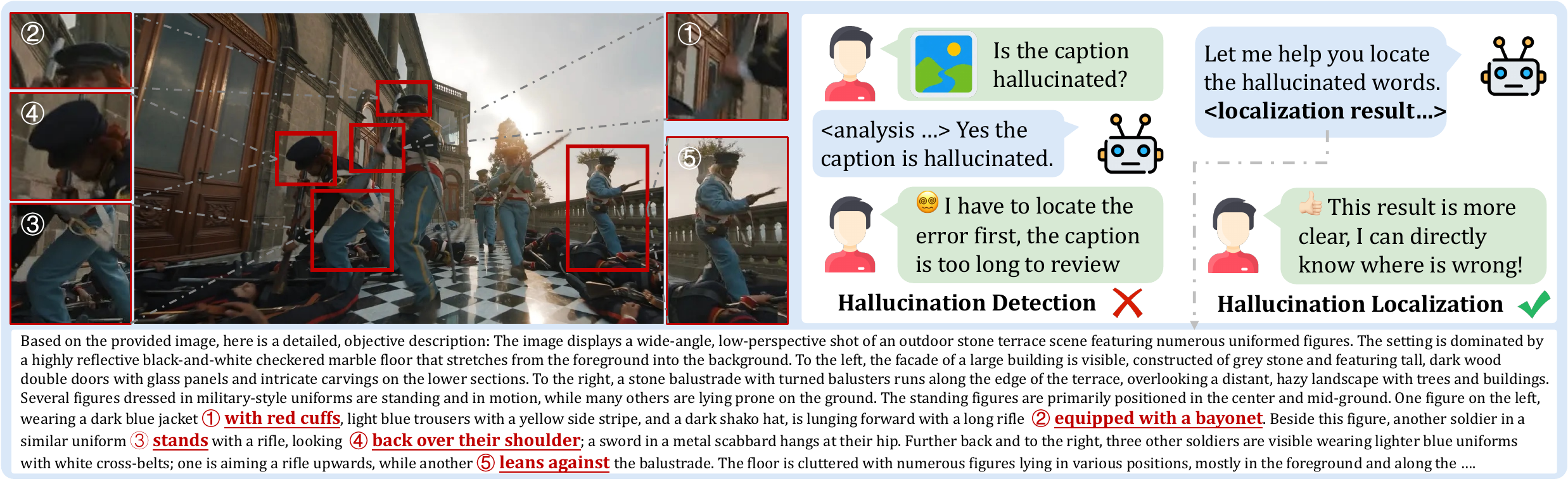}
  \caption{The difference between the task of hallucination detection and hallucination localization.}
  \label{fig:teaser}
\end{teaserfigure}


\maketitle
\section{Introduction}

The rapid advancement of Multimodal Large Language Models (MLLMs)~\cite{liu-nips-2023-v-ins-tune, gemini2-5-2025, qwen25vl-2025} has fundamentally transformed image captioning from producing brief, formulaic sentences into generating rich, paragraph-length narratives that describe visual content in exhaustive detail. Modern captioning systems~\cite{lu-2025-arxiv-omnicaptioner, xing-2025-arxiv-scalecap, xing-2025-arxiv-caprl} now routinely produce descriptions spanning hundreds of words, covering object attributes, spatial relationships, visual text, and other fine-grained dimensions that were previously beyond reach. This evolution toward ``hyper-detailed'' captioning has opened exciting possibilities for downstream applications such as fine-grained retrieval, visual generation and visual understanding.

However, this leap in descriptive capacity comes with a critical cost: as captions grow longer, the surface area for hallucination expands dramatically. Even a single subtle mistake---an incorrect object count, a misidentified color, or a fabricated spatial relationship---can undermine the trustworthiness of the entire narrative. As shown in Figure~\ref{fig:teaser}, it is not sufficient to merely determine \emph{whether} a caption contains errors; what users and downstream systems truly need is the ability to \emph{pinpoint exactly where} those errors occur in the caption, so that captions can be audited, corrected, and ultimately trusted.

Existing benchmarks, while valuable, fall short of evaluating models' ability to localize token-level hallucinations. Early benchmarks spent more effort on evaluating the quality of models' responses. For example, POPE~\cite{li-2023-emnlp-pope} evaluates hallucinations through binary polling queries on object existence. Later, some works focused on evaluating model responses at a more fine-grained level; for instance, FaithScore~\cite{jing-emnlp-2024-faithscore} decomposes atomic facts from the model response, and verifies the facts one by one. Unlike previous works that primarily emphasized interpretability, the task of hallucination localization requires models to pinpoint the exact locations of hallucinated words or sentences within a response. For example, ALOHa~\cite{petryk-2024-naacl-aloha} assesses object hallucination in the captions, and develops the HAllucination Test dataset to benchmark verifiers' ability to locate the object hallucinations. More recent works have begun to explore finer granularities: M-HalDetect~\cite{anisha-2024-aaai-mhaldet} provides segment-level annotations, HalLoc~\cite{park-2025-cvpr-halloc} introduces token-level hallucination localization, and TLDR~\cite{fu-2025-iclr-tldr} proposes a token-level reward model for self-correction. Yet these benchmarks share critical limitations: they operate on relatively short captions (typically under 60 words), cover only a single visual domain, and address a narrow taxonomy of hallucination types. As summarized in Table~\ref{tab:bmk-compare}, no existing benchmark simultaneously provides the caption length, annotation granularity, domain diversity, and hallucination taxonomy needed to rigorously evaluate MLLMs' dense hallucination localization ability in the era of long-form captioning.

To bridge this gap, we introduce \textbf{DetailVerifyBench}, the first benchmark specifically designed for \emph{dense hallucination localization} in hyper-detailed image captions. DetailVerifyBench comprises 1,000 high-quality images spanning five diverse domains---GUIs, natural scenes, charts, movie frames, and posters---with an average caption length exceeding 200 words. Each caption is annotated at the \emph{token level}, marking the precise boundaries of hallucinated content, and our taxonomy covers 10 fine-grained hallucination dimensions, such as object number, color, category, shape, material, camera, OCR. Specifically, we adopt a dual-source strategy: we not only utilize human annotators to annotate hallucinations in captions generated by Gemini-3-Pro to evaluate MLLMs on locating real hallucinations, but also provide a more cost-effective evaluation protocol via \emph{adversarial hallucination injection}. This pipeline uses an injector-detector adversarial loop to iteratively synthesize hard-to-detect errors and inject them directly into high-quality captions. Crucially, this injection approach eliminates the need for labor-intensive human annotation of errors and enables richer evaluation dimensions compared to real hallucinations.

Our contributions are summarized as follows:
\begin{itemize}
    \item \textbf{We present DetailVerifyBench}, a challenging benchmark for dense hallucination localization featuring 1,000 images across 5 domains, token-level annotations in captions averaging over 200 words, a 10-dimension hallucination taxonomy, and both real and synthetic hallucination variants.
    \item \textbf{We propose an adversarial hallucination injection} pipeline that generates hard-to-detect hallucinations through iterative injector-detector interaction, significantly elevating the benchmark's difficulty beyond previous injection methods.
    \item \textbf{We evaluate various advanced MLLMs} including open-source and closed-source commercial models. Evaluation results on multiple hallucination dimensions offer diagnostic insights into the strengths and failure modes of current MLLMs as hallucination localization verifiers.
\end{itemize}

\section{Related Works}

\subsection{MLLMs and Image Captioning} 
The evolution of image captioning has shifted from generating brief descriptions on datasets like MS-COCO \cite{lin-2015-arxiv-mscoco, chen-2015-arxiv-mscococap}, Conceptual Captions \cite{sharma-acl-2018-conceptual}, Local Narratives \cite{pont-2020-eccv-localnarratives} and Laion5B \cite{schuhmann-2022-nips-laion5b} to developing MLLMs \cite{liu-nips-2023-v-ins-tune,gemini2-5-2025,qwen25vl-2025,bai-2025-arxiv-qwen3vl,kimi-2026-arxiv-kimi2p5,stepfun-2025-arxiv-step3-vl-10b} capable of producing highly detailed, long-form descriptions. In order to better support the long captioning capabilities of MLLMs, more and more efforts are being devoted to creating long caption  datasets \cite{bonill-2024-arxiv-pixlore, li-2024-nips-densefusion, xiong-2024-nips-lvdm, xue-2025-nips-ultravideo}. For example, CapsFusion \cite{yu-2024-cvpr-capfusion} refines synthetic captions for better scalability, while DOCCI \cite{yasumasa-2024-arxiv-docci} provides long, human-annotated descriptions capturing spatial relations and fine details. Similarly, ImageInWords \cite{garg-2024-emnlp-imageinwords} introduces a human-in-the-loop framework for curating hyper-detailed annotations. Recent approaches like OmniCaptioner \cite{lu-2025-arxiv-omnicaptioner} unify captioning across diverse domains, while ScaleCap \cite{xing-2025-arxiv-scalecap} and CapRL \cite{xing-2025-arxiv-caprl} leverage inference-time scaling and reinforcement learning with verifiable rewards to enhance caption density and utility. Moreover, many high quality benchmarks \cite{liu-2025-neurips-capability, lu-2025-cvpr-comprecap,dong-2024-arxiv-capture, wang-2025-nips-cinetechbench} are designed to evaluate the quality of long captions from multiple angles (e.g., correctness and thoroughness). However, what remains underexplored is a capability that matters in practice: \textbf{can a model pinpoint the hallucination words in a long caption}---so that the description can be audited and, ultimately, corrected? As captions grow longer, even a single subtle error may undermine the reliability of the entire description, making error localization a critical step toward reliable long-form captioning.

\subsection{Hallucination Detection and Localization}
Visual hallucination in MLLMs primarily manifests as cross-modal inconsistency between generated textual outputs and visual inputs \cite{bai-2025-arxiv-hallsurvey, kaul-2024-cvpr-throne, qiu-2024-arxiv-longhalqa, chen-2024-acl-unihaldet, li-2023-emnlp-haelm, yan-2025-esa-bi}. Early efforts focused on detecting hallucinations in MLLM responses. For instance, POPE \cite{li-2023-emnlp-pope} utilized polling-based queries to evaluate object existence. Subsequent works like FaithScore \cite{jing-emnlp-2024-faithscore} and MOCHa \cite{ben-2024-emnlp-mocha} shifted toward verifying atomic facts and logical consistency. Building on this, the focus has increasingly transitioned from coarse-grained response-level classification to fine-grained hallucination localization. For instance, ALOHa \cite{petryk-2024-naacl-aloha} and HLVC \cite{nakada-2025-arxiv-videohalldet} assess specific hallucinated entities or events, successfully localizing them at the span level. M-HalDetect \cite{anisha-2024-aaai-mhaldet} further enriches this field with 16k segment-level annotations in image captions. Moving toward the finest granularity, the HalLoc \cite{park-2025-cvpr-halloc} dataset introduces token-level probabilistic detection across 155k samples. Beyond mere evaluation, such precise localization is now recognized as a vital prerequisite for hallucination mitigation; for example, TLDR \cite{fu-2025-iclr-tldr} leverages a token-level reward model to provide granular feedback for automated self-correction. Despite these advances in token-level granularity, existing datasets remain largely restricted to short descriptions within narrow domains. Consequently, in this era of long captioning, there is still a lack of a multi-domain benchmark to rigorously evaluate the ability of models to locate hallucinated words within the long captions.

\begin{table}[htbp]
\caption{\textbf{Comparison with existing hallucination localization benchmarks.} ``-'' means undisclosed. Gran.: annotation granularity; Real.: contains real hallucinations; Syn.: contains synthesized hallucinations.}
\label{tab:bmk-compare}
\centering
\footnotesize
\setlength{\tabcolsep}{3.5pt}
\begin{tabular}{l c c c c c c c}
\toprule
\textbf{Benchmark} & \textbf{Gran.} & \textbf{\#Samp.} & \textbf{Avg. Len.} & \textbf{\#Type} & \textbf{\#Dom.} & \textbf{Real.} & \textbf{Syn.} \\
\midrule
HaELM~\cite{li-2023-emnlp-haelm}            & Resp.  & 5{,}000   & $\sim$50 w & 2  & 1 &      & \checkmark \\
ALOHa~\cite{petryk-2024-naacl-aloha} & Span.  & 490  & $\sim$14 w & 1  & 1 & \checkmark  &  \\
M-HalDet~\cite{anisha-2024-aaai-mhaldet}     & Seg.   & 4{,}000   & $\sim$80 w & 3  & 1 & \checkmark     &  \\
HalLoc~\cite{park-2025-cvpr-halloc}          & Token  & 155{,}953 & $\sim$50 w & 3  & 1 &      & \checkmark \\
TLDR~\cite{fu-2025-iclr-tldr}                & Token  & --        & --         & -- & 1 &       & \checkmark \\
HLVC~\cite{nakada-2025-arxiv-videohalldet}    & Span   & 1{,}167   & $\sim$60 w & 3  & 1 & \checkmark     &  \\
\midrule
\textbf{Ours}                                 & \textbf{Token} & \textbf{1{,}000} & \textbf{$>$200 w} & \textbf{10} & \textbf{5} & \checkmark & \checkmark \\
\bottomrule
\end{tabular}
\end{table}

\section{Benchmark}

In this section, we first formulate the hallucination localization task and then introduce the building process of our benchmark. As summarized in Table~\ref{tab:bmk-compare}, our benchmark distinguishes itself from existing ones through three key features. First, it encompasses multiple domains, enabling a more comprehensive evaluation of caption verifiers. Second, it is the first open-source hallucination localization benchmark for long captions (averaging over 200 words) that provides token-level annotations. Third, it comprises two distinct versions: one containing real hallucinations from the advanced MLLM, Gemini-3-Pro, and the other featuring more diverse hallucination types introduced via our custom injection method.

\begin{table}[htbp]
    \caption{Information of each domain in our benchmark. Hallu. Rate: The ratio of captions contain hallucination.} 
    \centering
    \footnotesize
    \setlength{\tabcolsep}{3pt}
    \begin{tabular}{l l c c c c}
        \toprule
        \textbf{Domain} & \textbf{Source} & \textbf{\#Img} & \textbf{Avg. Len} & \textbf{\#Hallu locations} & \textbf{Hallu. Rate} \\
        \midrule
        GUI & Screenspot Pro \cite{li-2025-iclr-screenspotpro} & 200 & 196 & 274 & 68\% \\[2pt]
        Nature & DOCCI \cite{yasumasa-2024-arxiv-docci} & 200 & 148 & 69 & 26\% \\[2pt]
        Chart & Echarts Examples \footnote{https://echarts.apache.org/examples/en/index.html} & 200 & 197 & 192 & 41\% \\[2pt]
        \multirow{2}{*}{Movie} & CineTechBench \cite{wang-2025-nips-cinetechbench} & \multirow{2}{*}{200} & \multirow{2}{*}{214} & \multirow{2}{*}{613} & \multirow{2}{*}{88\%} \\
         & ShotBench \cite{liu-2025-nips-shotbench} & & & & \\[2pt]
        \multirow{2}{*}{Poster} & IMDB & \multirow{2}{*}{200} & \multirow{2}{*}{257} & \multirow{2}{*}{576} & \multirow{2}{*}{90\%} \\
         & Movie Poster 100k & & & & \\
        \bottomrule
    \end{tabular}
    \label{tab:bmk-domain-meta}
\end{table}

\subsection{Problem Formulation}
We cast hallucination localization as a constrained text generation problem. Formally, let $x$ denote an input image and let $c = (c_1, c_2, \dots, c_N)$ a candidate caption consisting of $N$ tokens. The localization model $\pi_\theta$ is tasked with generating an augmented output sequence $o$ that satisfies two constraints simultaneously:

\paragraph{\textbf{Lexical Faithfulness.}} The output $o$ must faithfully reproduce every token in $c$, preserving the original word order and content. That is, after stripping all annotation tags from $o$, the resulting plain-text sequence must be identical to $c$.

\paragraph{\textbf{Hallucination Localization.}} The model must find all hallucinated tokens and wrap them in boundary tags \texttt{<HALLUCINATION>} and \texttt{</HALLUCINATION>}. Let $\mathcal{H} \subseteq \{1, 2, \dots, N\}$ denote the set of hallucinated token indices in the ground truth. From the augmented output $o$, we extract the predicted index set $\hat{\mathcal{H}}$, and performance is measured by token-level Precision, Recall, and F1:
\begin{equation}
    P = \frac{|\hat{\mathcal{H}} \cap \mathcal{H}|}{|\hat{\mathcal{H}}|}, \quad
    R = \frac{|\hat{\mathcal{H}} \cap \mathcal{H}|}{|\mathcal{H}|}, \quad
    F_1 = \frac{2PR}{P + R}.
    \label{eq:token-metrics}
\end{equation}

For sentence-level evaluation, a sentence $s_j$ is considered hallucinated if it contains at least one hallucinated token, i.e., $\mathcal{H} \cap \mathcal{I}(s_j) \neq \emptyset$, where $\mathcal{I}(s_j)$ is the index set of tokens in $s_j$. Similarly, a predicted sentence is labeled as hallucinated if $\hat{\mathcal{H}} \cap \mathcal{I}(s_j) \neq \emptyset$. Precision, Recall, and F1 are then computed over sentence-level binary labels.

\subsection{Benchmark Building Pipeline}

\begin{figure}[htbp]
    \centering
    \includegraphics[width=\linewidth]{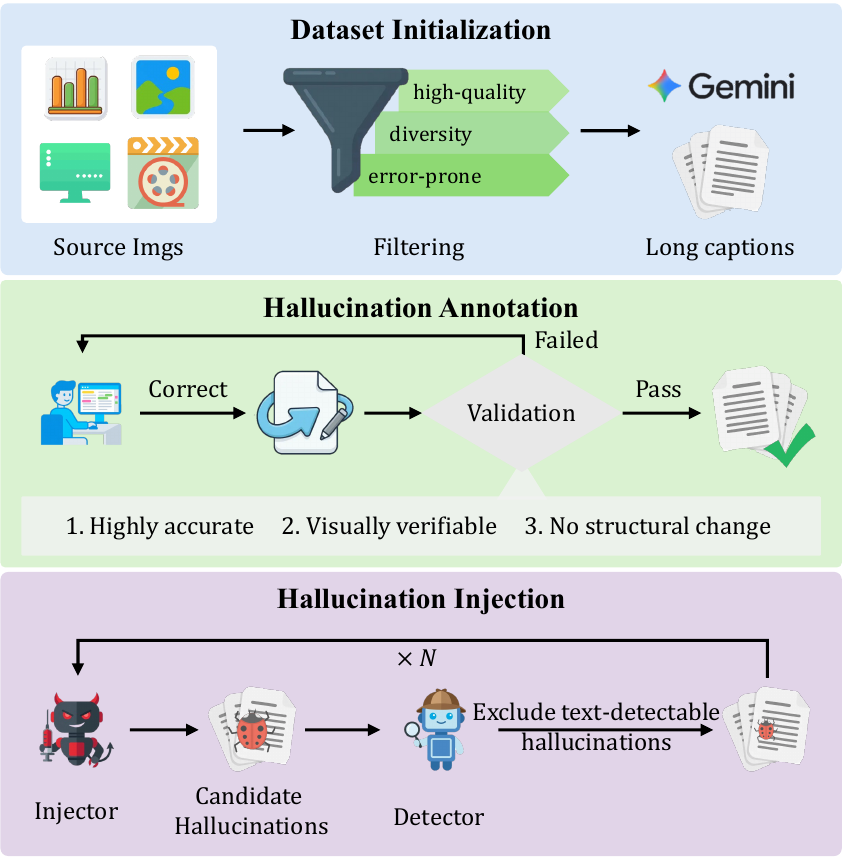}
    \caption{The pipeline for building the DetailVerifyBench.}
    \label{fig:data-pipe}
\end{figure}

Figure~\ref{fig:data-pipe} shows three stages in our benchmark building process:
\begin{itemize}
    \item \textbf{Stage 1: Dataset Initialization.} First, we carefully select high-quality, hallucination-prone images across diverse domains (e.g., GUI, Nature, Chart, Movie, and Poster) to form the foundation of our benchmark. Detailed information is provided in the Table \ref{tab:bmk-domain-meta}. Then we utilize Gemini-3-Pro to generate initial, domain-specific long descriptions. All domain prompts are specifically designed to request the description of verifiable visual facts rather than subjective interpretations—such as emotional atmosphere or predictions of future narrative events. This provides high-quality caption drafts, reducing the "cold-start" burden on human annotators.
    
    \item \textbf{Stage 2: Hallucination Annotation.} This stage involves hallucination extraction via human correction. Annotators correct specific visual fact errors (e.g., object counts, attributes, or relationships) to obtain "clean" texts, automatically yielding token-level hallucination boundaries via text diff. A key constraint here is to maintain the original sentence structure and vocabulary of the MLLM output. To guarantee annotation quality, the annotators are required to do cross check. In the final acceptance stage, we implement a batch-based verification protocol with a strict acceptance threshold of 97\%; any batch failing to meet this standard is returned for revision. 

    \item \textbf{Stage 3: Hallucination Injection.} To further challenge localization models, we employ an adversarial injection method to introduce hard-to-detect hallucinations into the clean captions. We will elaborate on this injection methodology in the subsequent section.

\end{itemize}

\subsection{Hallucination Injection}

Following the acquisition of verified ground truth captions, we synthesize hallucinated captions as negative samples. A straightforward approach, similar to TLDR~\cite{fu-2025-iclr-tldr}, employs LLMs to perturb visual facts directly within the text. However, since the perturbation model lacks access to the actual image, the resulting hallucinations are often physically implausible or trivially detectable by language priors alone. To ensure injected hallucinations are genuine ``hard negatives'' that require visual verification, we propose an \emph{adversarial injection} pipeline. Our method iteratively refines injected hallucinations through a two-player adversarial loop between an \textbf{injector} and a \textbf{detector}:

\begin{table*}[t]
\centering
\caption{Evaluation results of advanced open-source and closed-source MLLMs on real and synthetic hallucinations. Dimension abbreviations: Num (Number), Clr (Color), Cat (Category), Shp (Shape), Mat (Material), Spat (Spatial), OCR (Optical Character Recognition), Scene (Scene), Cam (Camera). The best results are in bold type.}
\label{tab:main_results}
\footnotesize
\setlength{\tabcolsep}{2.5pt}

\begin{tabular}{l|ccc|cccccccc|ccc|cccccccccc}
\toprule

\multirow{2}{*}{Model} & \multicolumn{3}{c|}{\textbf{Real Overall}} & \multicolumn{8}{c|}{\textbf{Real Dimension-$R_{tok}$}}
 & \multicolumn{3}{c|}{\textbf{Synthetic  Overall}} & \multicolumn{10}{c}{\textbf{Synthetic Dimension-$R_{tok}$}} \\

 & \small $\mathrm{P_{tok}}$ & \small $\mathrm{R_{tok}}$ & \small $\mathrm{F1_{tok}}$
 & \scriptsize Num & \scriptsize Clr & \scriptsize Cat & \scriptsize Shp & \scriptsize Mat & \scriptsize Spat & \scriptsize OCR & \scriptsize Other
 & \small $\mathrm{P_{tok}}$ & \small $\mathrm{R_{tok}}$ & \small $\mathrm{F1_{tok}}$
 & \scriptsize Num & \scriptsize Clr & \scriptsize Cat & \scriptsize Shp & \scriptsize Mat & \scriptsize Spat & \scriptsize OCR & \scriptsize Scene & \scriptsize Cam & \scriptsize Other \\

\midrule

\textit{Open-source MLLMs} & & & & & & &  & & & & & & & \\
GLM-4.6V-Flash        & 0.01 & 0.01 & 0.01 & 0.00 & 0.00 & 0.01 & 0.01 & 0.00 & 0.00 & 0.01 & 0.01 & 0.41 & 0.22 & 0.26 & 0.17 & 0.40 & 0.27 & 0.17 & 0.40 & 0.06 & 0.34 & 0.31 & 0.00 & 0.22 \\
Step3-VL-10B           & 0.03 & 0.04 & 0.03 & 0.03 & 0.04 & 0.13 & 0.07 & 0.13 & 0.04 & 0.15 & 0.04 & 0.36 & 0.48 & 0.35 & 0.49 & 0.66 & 0.48 & 0.43 & 0.53 & 0.37 & 0.77 & 0.68 & 0.04 & 0.39 \\
Qwen3-VL-8B-Thinking  & 0.04 & 0.03 & 0.03 & 0.02 & 0.04 & 0.03 & 0.15 & 0.02 & 0.01 & 0.10 & 0.03 & 0.45 & 0.47 & 0.39 & 0.45 & 0.68 & 0.52 & 0.41 & 0.63 & 0.25 & 0.76 & 0.64 & 0.05 & 0.49 \\
Qwen3.5-9B            & 0.10 & 0.09 & 0.08 & 0.16 & 0.10 & 0.11 & 0.12 & 0.28 & 0.05 & 0.24 & 0.11 & 0.66 & 0.60 & 0.58 & 0.49 & 0.74 & 0.58 & 0.48 & 0.61 & 0.50 & 0.86 & 0.61 & 0.22 & 0.63 \\
Qwen3.5-35B-A3B       & 0.12 & 0.11 & 0.10 & 0.18 & 0.16 & 0.12 & 0.13 & 0.26 & 0.09 & 0.34 & 0.13 & 0.55 & 0.64 & 0.52 & 0.55 & 0.75 & 0.60 & 0.53 & 0.56 & 0.60 & 0.86 & 0.68 & 0.15 & 0.57 \\
Qwen3.5-397B-A17B     & 0.15 & \textbf{0.16} & \textbf{0.13} & \textbf{0.26} & \textbf{0.22} & 0.22 & \textbf{0.21} & 0.37 & 0.14 & 0.38 & \textbf{0.21} & 0.62 & \textbf{0.70} & 0.61 & 0.65 & \textbf{0.79} & \textbf{0.65} & 0.57 & 0.67 & 0.69 & \textbf{0.89} & 0.69 & 0.22 & \textbf{0.69} \\
KIMI-K2.5              & 0.11 & 0.08 & 0.08 & 0.05 & 0.09 & 0.09 & 0.11 & 0.10 & 0.07 & 0.28 & 0.09 & 0.69 & 0.57 & 0.58 & 0.52 & 0.67 & 0.52 & 0.43 & 0.50 & 0.56 & 0.80 & 0.63 & 0.13 & 0.52 \\

\midrule
\textit{MLLMs + VCD} & & & & & & &  & & & & & & & \\
Qwen3-VL-8B-Thinking  & 0.02 & 0.02 & 0.02 & 0.02 & 0.04 & 0.05 & 0.01 & 0.03 & 0.01 & 0.01 & 0.03 & 0.25 & 0.09 & 0.12 & 0.03 & 0.13 & 0.07 & 0.10 & 0.17 & 0.03 & 0.04 & 0.14 & 0.07 & 0.10 \\

\midrule
\textit{Closed-source MLLMs} & & & & & & &  & & & & & & & \\
Seed2.0-pro            & 0.13 & 0.12 & 0.11 & 0.21 & 0.14 & \textbf{0.23} & 0.10 & \textbf{0.46} & \textbf{0.16} & 0.35 & 0.20 & 0.59 & 0.68 & 0.57 & \textbf{0.65} & 0.77 & 0.63 & 0.53 & \textbf{0.68} & 0.66 & 0.86 & \textbf{0.77} & \textbf{0.33} & 0.64 \\
Mimo-v2-pro            & 0.01 & 0.01 & 0.01 & 0.03 & 0.01 & 0.00 & 0.01 & 0.00 & 0.01 & 0.01 & 0.00 & 0.06 & 0.03 & 0.03 & 0.05 & 0.03 & 0.03 & 0.00 & 0.02 & 0.01 & 0.03 & 0.05 & 0.00 & 0.08 \\
GPT-5.2                & 0.06 & 0.08 & 0.05 & 0.09 & 0.07 & 0.10 & 0.07 & 0.19 & 0.09 & 0.27 & 0.12 & 0.46 & 0.53 & 0.43 & 0.57 & 0.64 & 0.51 & 0.46 & 0.58 & 0.41 & 0.82 & 0.75 & 0.14 & 0.58 \\
GPT-5.4                & 0.07 & 0.09 & 0.07 & 0.13 & 0.07 & 0.11 & 0.11 & 0.09 & 0.12 & 0.26 & 0.12 & 0.43 & 0.61 & 0.45 & 0.63 & 0.70 & 0.58 & 0.57 & 0.63 & 0.56 & 0.86 & 0.74 & 0.14 & 0.65 \\
Gemini-3-Pro-Preview    & 0.13 & 0.10 & 0.10 & 0.12 & 0.10 & 0.13 & 0.13 & 0.17 & 0.09 & \textbf{0.40} & 0.10 & 0.70 & 0.66 & 0.64 & 0.55 & 0.72 & 0.61 & 0.57 & 0.66 & 0.63 & 0.88 & 0.68 & 0.23 & 0.67 \\
Gemini-3.1-Pro-Preview & \textbf{0.16} & 0.12 & 0.12 & 0.17 & 0.11 & 0.17 & \textbf{0.21} & 0.22 & 0.13 & 0.37 & 0.15 & \textbf{0.76} & 0.66 & \textbf{0.66} & 0.59 & 0.73 & 0.63 & \textbf{0.62} & 0.57 & 0.62 & 0.87 & 0.69 & 0.17 & 0.66 \\
Claude-Opus-4.6        & 0.09 & 0.07 & 0.07 & 0.06 & 0.08 & 0.12 & 0.06 & 0.03 & 0.06 & 0.30 & 0.08 & 0.22 & 0.07 & 0.09 & 0.09 & 0.06 & 0.07 & 0.07 & 0.05 & 0.06 & 0.10 & 0.05 & 0.05 & 0.10 \\

\bottomrule
\end{tabular}
\end{table*}

\paragraph{\textbf{Injection.}} Given the clean caption $c$, the injector LLM (here we use Gemini-3-Flash) modifies specific visual facts in $c$ to produce a hallucinated caption $c'$. Specifically, the injector is designed to target ten distinct hallucination categories: number, color, category, shape, material, spatial relation, scene, camera, OCR, and other (for errors that do not fall into the aforementioned types). To ensure the introduced errors are both semantically coherent and highly deceptive, the injector is prompted to follow a structured, four-step generation strategy: First, it identifies error-prone words and their specific locations within the original description. Second, it proposes candidate replacements that maintain contextual fluency and physical plausibility, ensuring the errors are grounded in concrete visual elements. Third, it selects the optimal modification for each location. Finally, it formats the output $c'$ by explicitly wrapping only the modified spans within \texttt{<HALLUCINATION>} tags.

\paragraph{\textbf{Detection.}} A separate detector LLM (here we use GPT-5.2), which receives \emph{only} the hallucinated caption $c'$ \emph{without} the image and hallucination tags, attempts to identify the injected hallucinations based solely on textual cues (e.g., internal inconsistencies, implausible descriptions, or common-sense violations).

\begin{figure*}[tbp]
    \centering
    \includegraphics[width=\textwidth]{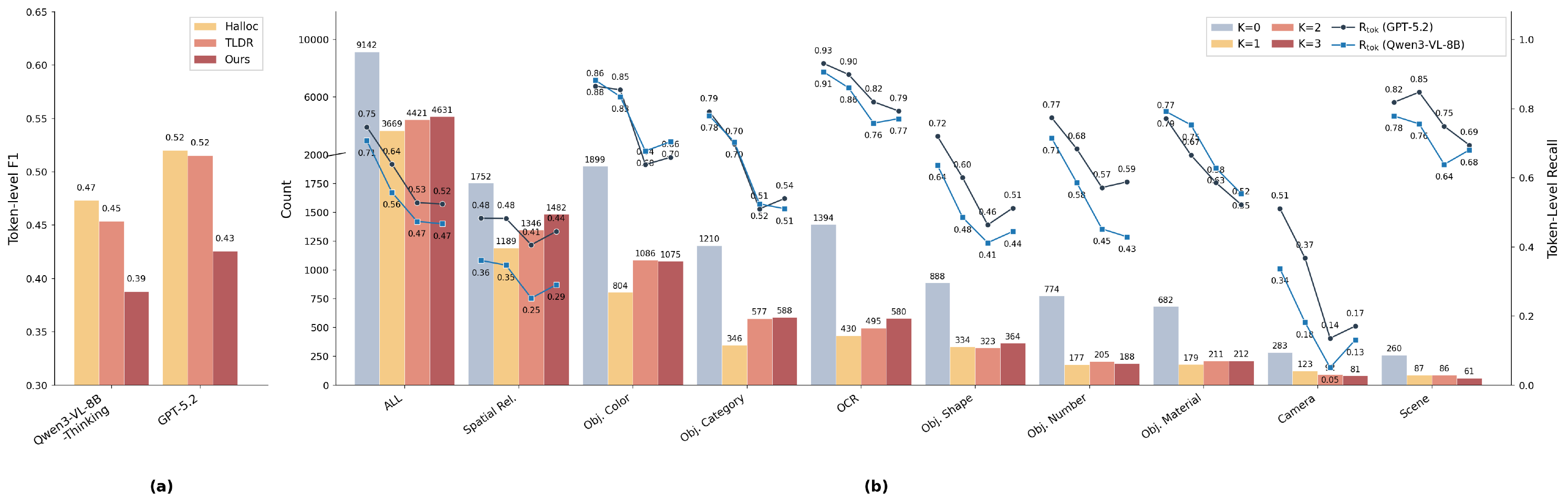}
    \caption{(a) Comparison of injection methods; (b) Hallucination counts \& $\mathrm{R_{tok}}$ of different adversarial iteration $K$.}
    \label{fig:k-ablation}
\end{figure*}

\paragraph{\textbf{Filtering and iterating with feedback.}} Hallucinations successfully detected by the text-only detector are removed from $c'$, as they are considered ``easy''—identifiable without visual grounding. The detection results, including which spans were caught and why, are fed back to the injector to guide the next injection iteration. This loop repeats for $K$ iterations. The surviving hallucinations after the final round are those that \emph{cannot} be identified without visual evidence, ensuring the benchmark genuinely tests visual grounding capability rather than linguistic plausibility. We set $K=2$ based on the ablation results observed in Figure~\ref{fig:k-ablation}.

\begin{figure}[htbp]
    \centering
    \includegraphics[width=0.5\textwidth]{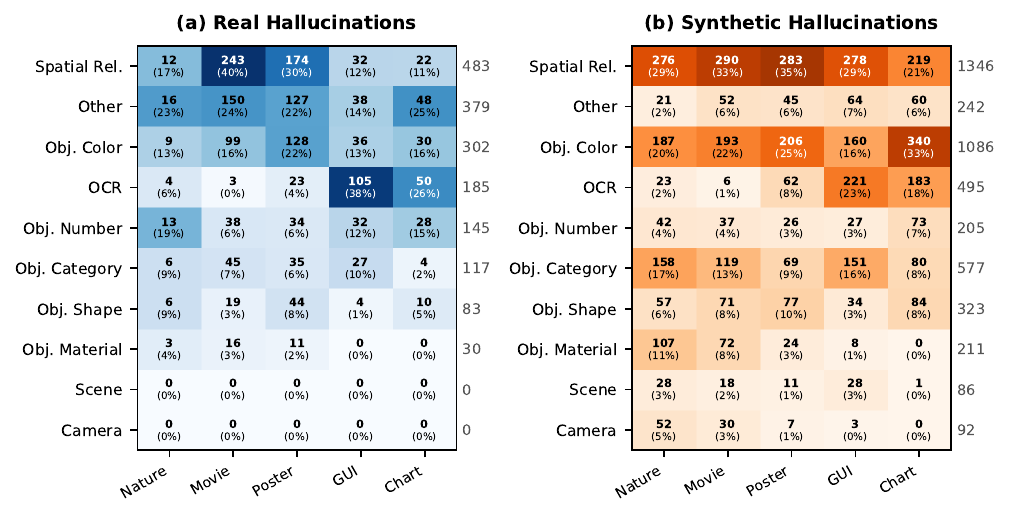}
    \caption{Distribution of 10 hallucination dimensions across five domains for (a) real and (b) synthetic hallucinations. Each cell shows the count (top) and percentage within the domain (bottom). The rightmost column indicates the total count per dimension. Color intensity reflects the within-domain percentage.}
    \label{fig:hallu-analysis}
\end{figure}

\subsection{Benchmark Meta Information}

\paragraph{\textbf{Hallucination distribution}} Figure~\ref{fig:hallu-analysis} compares the hallucination distribution across the real and synthetic benchmark versions. The real set exhibits sparse coverage in Scene (8) and Camera (1), whereas the synthetic set fills these gaps (86 and 92, respectively), ensuring all 10 dimensions are adequately tested. Meanwhile, real hallucination profiles are strongly domain-dependent: Movie and Poster domains are dominated by spatial relation hallucinations, GUI and Chart by OCR errors, and Nature by Object Color and Category. 

\paragraph{\textbf{Annotation cost}} We outsourced the data annotation task to a professional annotation company. For lengthy image captions, pricing is determined by two scenarios: If the caption generated by Gemini-3-Pro contains hallucinations, the unit price is set at 17 CNY. If the caption is accurate, the unit price is 12 CNY. The higher cost for identifying hallucinations reflects the substantial cognitive load required to carefully compare lengthy captions with the corresponding images. To incentivize thorough checking, we have assigned a higher unit price for annotating the hallucinated captions. The complete dataset is publicly released under the CC-BY-4.0 license. For further annotation details, please visit our website.

\section{Experiments}
\label{sec:experiments}

\subsection{Evaluation Settings}

We evaluate more than ten various state-of-the-art MLLMs spanning both open-source and closed-source families. For open-source models, we include GLM-4.6V-Flash \cite{GLM-2026-arxiv-GLM-4.5V}, Step3-VL-10B \cite{stepfun-2025-arxiv-step3-vl-10b}, Qwen3-VL-8B-Thinking \cite{bai-2025-arxiv-qwen3vl}, Qwen3.5 series (9B, 35B-A3B, 397B-A17B) \cite{noauthor_qwen_nodate}, KIMI-K2.5 \cite{kimi-2026-arxiv-kimi2p5}. We also evaluate a decoding-based hallucination mitigation plugin, visual contrastive decoding~\cite{leng-2024-cvpr-vcd}, applied on top of Qwen3-VL-8B-Thinking. For closed-source models, we test Seed2.0-pro \cite{seed2.0}, Mimo-v2-pro \cite{mimov2-flash-technical-report}, GPT-5.2 \cite{gpt-5-2}, GPT-5.4 \cite{gpt-5-4}, Gemini-3-Pro-Preview, Gemini-3.1-Pro-Preview \cite{gemini3}, and Claude-Opus-4.6. All models are prompted with a unified instruction that requires them to reproduce the input caption while wrapping hallucinated tokens with \texttt{<HALLUCINATION>}\texttt{</HALLUCINATION>} tags. For overall evaluation, we adopt Precision ($P$), Recall ($R$), and F1-score ($F_1$) as core metrics, computed at granularities: \emph{token level} and \emph{sentence level}. Beyond overall performance, we report \emph{dimension-level token recall} ($R_{\mathrm{tok}}$) for each hallucination category in our taxonomy.

\vspace{-0.24cm}

\subsection{Main Results}

\begin{figure}[htbp]
    \centering
    \includegraphics[width=\linewidth]{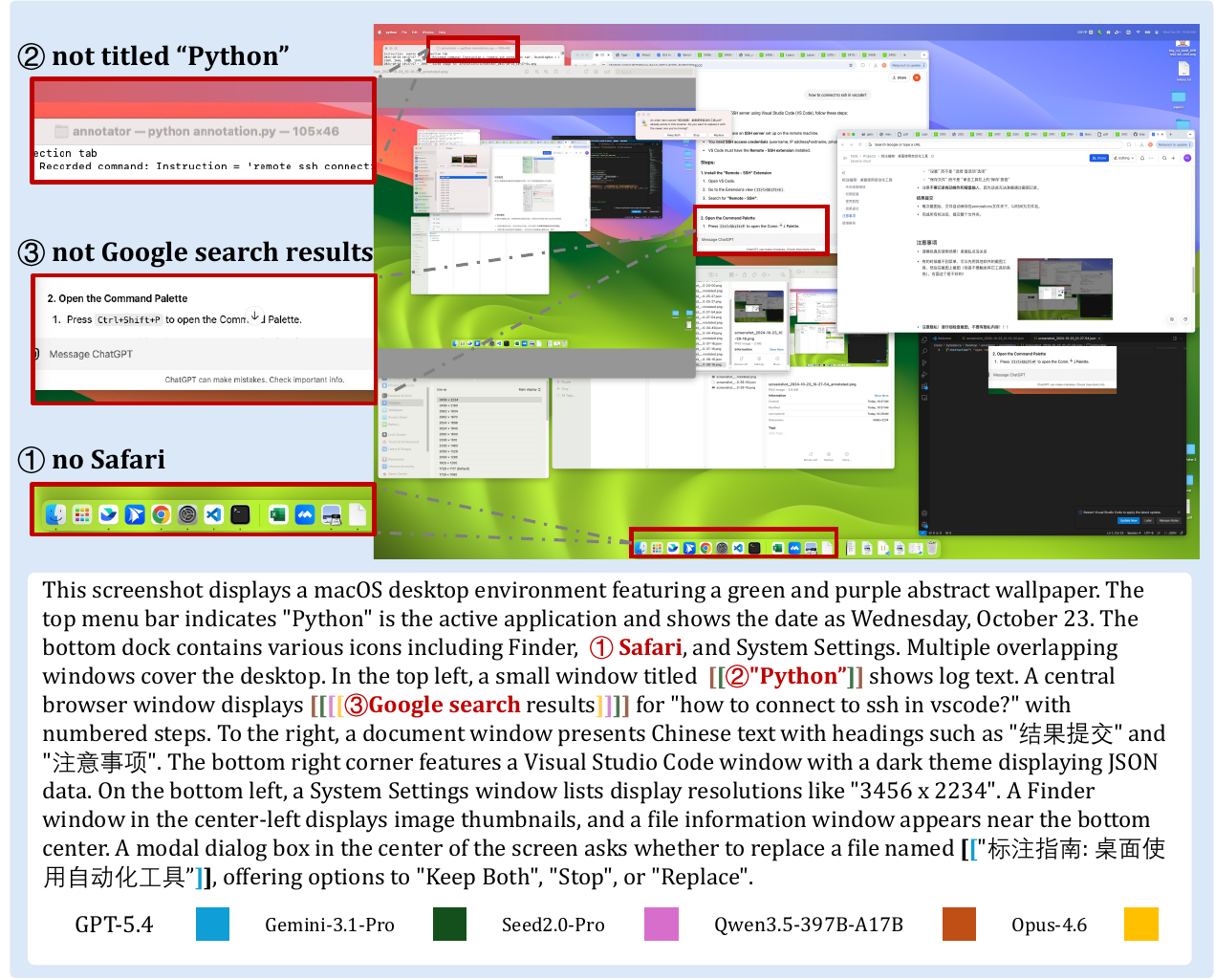}
    \caption{Visualization of a GUI image example with real hallucinations in our dataset. Tiny areas were enlarged and the judgments were annotated beside. The red bold text is the ground truth localization annotation. [ ] with specific color indicates the localization result of each MLLM. For more visualization examples, please visit our website.}
    \label{fig:vis-example}
\end{figure}

\paragraph{\textbf{Results on real hallucinations.}}
Table~\ref{tab:main_results} presents the evaluation results on real hallucinations. There are several key observations: \textbf{(1) Dense hallucination localization remains extremely challenging.} Even one of the most advanced MLLMs, Gemini-3.1-Pro-Preview, achieves only 0.12 token-level F1, underscoring the difficulty of precisely localizing hallucinated spans in long-form captions. Due to the high difficulty, there is no clear performance gap between closed-source and open-source models. By token-level F1, Qwen3.5-397B-A17B performs best, with 0.15 token-level precision and 0.16 recall, while Gemini-3.1-Pro-Preview achieves the highest precision (0.16). \textbf{(2) Models exhibit distinct dimensional biases.} OCR hallucinations are relatively easier to locate across most MLLMs, with Gemini-3-Pro-Preview achieving 0.40 dimension recall on OCR. In contrast, spatial relation hallucinations—the most prevalent category in our benchmark—prove consistently difficult, with most models scoring below 0.15. Intra-family models exhibit the same dimensional bias, for example, Qwen3.5 series (9B to 397B) consistently good at locating material and OCR hallucinations. \textbf{(3) Hallucinations induced by common environmental contexts are particularly difficult to detect.} As shown in Figure~\ref{fig:vis-example}, the Safari icon is a standard component of the macOS dock. Due to this strong prior, most MLLMs struggle to locate when its presence is hallucinated.

\paragraph{\textbf{Results on synthetic hallucinations.}}
As shown in the right half of Table~\ref{tab:main_results}, two key findings emerge.
\textbf{(1) Synthetic hallucinations are substantially easier to detect.} Gemini-3-Pro-Preview achieves 0.64 F1$_\mathrm{tok}$ on synthetic data versus 0.10 on real data, and GPT-5.4 reaches 0.45 versus 0.07. This gap is expected: injected hallucinations are localized perturbations to clean text, whereas real hallucinations are deeply entangled with the generation process.
\textbf{(2) Model rankings are highly consistent across the two settings.} Spearman's rank correlation on token-level metrics yields $\rho{=}0.838$ (P$_\mathrm{tok}$), $0.947$ (R$_\mathrm{tok}$), and $0.856$ (F1$_\mathrm{tok}$), all with $p{<}0.001$. This suggests that synthetic evaluation can reflect real-world relative performance, supporting its use for scalable, automated benchmarking.

\subsection{Ablation}

\paragraph{\textbf{Hallucination injection methods.}}
We compare our adversarial injection pipeline against alternative hallucination synthesis methods from previous works. All methods are evaluated using the same MLLMs (Qwen3-VL-8B-Thinking and GPT-5.2). Since our goal is to produce hard-to-detect hallucinations that require visual verification, \emph{lower} MLLMs' locating performance indicates higher injection performance. As shown in Figure~\ref{fig:k-ablation}(a), compared to the injection methods in HalLoc and TLDR, our adversarial injection method achieves lower F1.
The visualized injection examples from different methods are shown on our website.

\paragraph{\textbf{Adversarial Iteration Rounds $K$.}}
We investigate how the number of adversarial iterations $K$ between the injector and detector affects the quality of injected hallucinations. Table~\ref{tab:adv-iterations} and Figure~\ref{fig:k-ablation} reveal consistent trends across both detector models. At $K=0$ (single-pass injection without adversarial filtering), both detectors achieve high recall (0.75 for GPT-5.2 and 0.71 for Qwen3-VL-8B) and F1 (0.63 and 0.57), indicating that a substantial portion of initial injections are easily detectable. After one adversarial round ($K=1$), performance drops sharply: F1 decreases to 0.48 and 0.42 respectively, with recall falling from 0.75 to 0.64 for GPT-5.2 and from 0.71 to 0.56 for Qwen3-VL-8B, demonstrating effective removal of textually detectable hallucinations. A second round ($K=2$) yields further improvement, reducing F1 to 0.43/0.39. However, moving to $K=3$, performance plateaus at nearly identical levels (F1 remains 0.43/0.39), suggesting the adversarial loop has converged. As shown in Figure~\ref{fig:k-ablation}, this convergence pattern is consistent across individual hallucination dimensions: the largest drops occur from $K=0$ to $K=2$, with diminishing returns thereafter. We therefore adopt $K=2$ as our default setting, balancing injection difficulty and computational efficiency.

\begin{table}[htbp]
\centering
\footnotesize
\caption{\textbf{Adversarial injection quality across iteration rounds.} We report the token-level precision ($\mathrm{P}_{\mathrm{tok}}$), recall ($\mathrm{R}_{\mathrm{tok}}$), and F1 score ($\mathrm{F}_{\mathrm{tok}}$) for different models (GPT-5.2 and Qwen3-VL-8B-Thinking), measured at each iteration round $K$. Lower values indicate harder hallucinations.}
\label{tab:adv-iterations}
\begin{tabular}{@{}ll cccc@{}}
\toprule
\multirow{2}{*}{\textbf{Model}} & \multirow{2}{*}{\textbf{Metric}} & \multicolumn{4}{c}{\textbf{Iteration Round $K$}} \\
\cmidrule(lr){3-6}
& & \textbf{0} & \textbf{1} & \textbf{2} & \textbf{3} \\
\midrule
\multirow{3}{*}{GPT-5.2} 
& $\mathrm{P}_{\mathrm{tok}}$ & 0.60 & 0.48 (-0.12) & 0.46 (-0.14) & 0.46 (-0.14) \\
& $\mathrm{R}_{\mathrm{tok}}$ & 0.75 & 0.64 (-0.11) & 0.53 (-0.22) & 0.52 (-0.23) \\
& $\mathrm{F}_{\mathrm{tok}}$ & 0.63 & 0.48 (-0.15) & 0.43 (-0.20) & 0.43 (-0.20) \\
\midrule
\multirow{3}{*}{Qwen3-VL-8B-Thinking} 
& $\mathrm{P}_{\mathrm{tok}}$ & 0.53 & 0.45 (-0.08) & 0.45 (-0.08) & 0.44 (-0.09) \\
& $\mathrm{R}_{\mathrm{tok}}$ & 0.71 & 0.56 (-0.15) & 0.47 (-0.24) & 0.47 (-0.24) \\
& $\mathrm{F}_{\mathrm{tok}}$ & 0.57 & 0.42 (-0.15) & 0.39 (-0.18) & 0.39 (-0.18) \\
\bottomrule
\end{tabular}
\end{table}

\section{Conclusion}

We present DetailVerifyBench, a benchmark for dense hallucination localization in long image captions, featuring 1,000 images across 5 domains with token-level annotations averaging over 200 words. Our adversarial injection pipeline generates hard-negative hallucinations that require visual inspection to detect. Extensive evaluation reveals that even the strongest MLLMs achieve modest localization performance, highlighting significant room for improvement. However, a fidelity gap remains between synthetic and real hallucinations. Future research should focus on bridging this disparity to ensure that synthetic data can more faithfully mimic real-world hallucination patterns. We hope DetailVerifyBench will catalyze future research on fine-grained hallucination localization and reliable long-form captioning.


\bibliographystyle{ACM-Reference-Format}
\bibliography{sample-base}

\appendix

\end{document}